\documentclass[conference]{IEEEtran}
\IEEEoverridecommandlockouts
\usepackage{cite}
\usepackage{amsmath,amssymb,amsfonts}
\usepackage{graphicx}
\usepackage{textcomp}
\usepackage{xcolor}

\usepackage{enumitem}
\usepackage{algorithmicx}
\usepackage{algorithm} 
\usepackage{threeparttable}
\usepackage{multirow}
\usepackage{subfigure}
\usepackage{booktabs}
\usepackage[noend]{algpseudocode}
\usepackage{hyperref}

\def\BibTeX{{\rm B\kern-.05em{\sc i\kern-.025em b}\kern-.08em
    T\kern-.1667em\lower.7ex\hbox{E}\kern-.125emX}}
    
\newcommand{\sysName}{Purified Learning}

\begin{document}

\title{Learning Purified Feature Representations \\ from Task-irrelevant Labels
}

\author{\IEEEauthorblockN{Yinghui Li$^{1, *}$\thanks{$^{*}$ indicates equal contribution.}, Chen Wang$^{2, *}$, Yangning Li$^{1, *}$, Hai-Tao Zheng$^{1,4,\dagger}$\thanks{ $^{\dagger}$ Corresponding author: Hai-Tao Zheng, Ying Shen. (E-mail: zheng.haitao@sz.tsinghua.edu.cn, sheny76@mail.sysu.edu.cn)}, Ying Shen$^{3,\dagger}$}
\IEEEauthorblockA{\textit{$^{1}$ Tsinghua Shenzhen International Graduate School, Tsinghua University} \\
\textit{$^{2}$Department of Computer Science and Technology, Tsinghua University}\\
\textit{$^{3}$School of Intelligent Systems Engineering, Sun-Yat Sen University}\\
\textit{$^{4}$Peng Cheng Laboratory} \\
\{liyinghu20, wchen20, liyn20\}@mails.tsinghua.edu.cn}

}
% \author{\IEEEauthorblockN{Anonymous Authors}}

\maketitle

\begin{abstract}
Learning an empirically effective model with generalization using limited data is a challenging task for deep neural networks. In this paper, we propose a novel learning framework called \emph{\sysName{}} to exploit task-irrelevant features extracted from task-irrelevant labels when training models on small-scale datasets. Particularly, we purify feature representations by using the expression of task-irrelevant information, thus facilitating the learning process of classification. 
Our work is built on solid theoretical analysis and extensive experiments, which demonstrate the effectiveness of \sysName{}. 
According to the theory we proved, \sysName{} is model-agnostic and doesn't have any restrictions on the model needed, so it can be combined with any existing deep neural networks with ease to achieve better performance.
% The source code of this paper will be available in the future for reproducibility.
\end{abstract}


\section{Introduction}
With sufficient manually annotated training samples, deep neural networks could automatically perform feature extraction and achieve unprecedented performances on various classification tasks  \cite{gu2018recent}. 
However, collecting and annotating adequate training data is an extremely time-consuming and expensive process, leading that in many instances the training samples are insufficient and even noisy  \cite{4}. Under this circumstance, the performance of deep models always drops gravely on most classification tasks \cite{1,2}. 
The essential reason for the phenomenon is that the goal of ``minimizing the empirical risk" is not reliable when the training data is inadequate \cite{3}, as a result, deep neural models will easily overfit the training data. Therefore, training effective deep neural models with remarkable generalization performance on small training samples is of great practical importance in terms of vastly expanding the scalability of deep learning methods.

Many techniques have been developed to tackle the issue. Some approaches directly expanding training samples  \cite{cubuk2018autoaugment,bowles2018gan,33,13,27} to alleviate the shortage of annotated data, but may be limited because expansion based on small-scale training samples is only of theoretical significance and not operability.
On the other hand, previous studies also attempt to break through the limitation of a specific dataset and find training samples in a broad sense. Based on this motivation, transfer learning \cite{10,12} exploits knowledge from additional datasets with relevant content and labels, achieving remarkable improvements on the target task. 
However, they only achieve good results when labels are relevant enough since irrelevant labels could bring out massive negative transfer \cite{15,16}.
And to date, there have been few studies that use task-irrelevant features to improve the performance of deep learning models.
In this paper, we argue that while transferring task-relevant labels from other datasets, task-irrelevant labels could also be utilized to improve the generalization of classification on small datasets without conflicts.

Consider a running example in computer vision: facial expression recognition a.k.a FER, which aims at constructing a model to accurately predict the expression of unseen facial pictures. 
In this task, it is obvious that \emph{smiling} is a task-relevant feature, while \emph{hair color} is a proper task-irrelevant one.
Assume we have a small-scale FER dataset, which is often the case in real-world deep learning tasks. Since the training dataset is extremely small, it is unavoidable that the distribution of task-irrelevant features i.e., \emph{hair color} have non-negligible bias. For example, most people with black hair might be labeled as \emph{happy} and people with white hair are exactly labeled as \emph{sad}.
This kind of bias is a ubiquitous problem when training samples are insufficient, which has already been observed in many previous works  \cite{tommasi2017deeper,torralba2011unbiased}.

As we mentioned before, the task-irrelevant features create a substantial barrier for the learning process due to the problem of ``minimizing the empirical risk", which leads to the models inevitably overfit hair features. As a result, the performance of the trained models on unseen facial expression samples are severely affected. 
An intuitive method to address the issue is to make use of massive related facial information from other facial datasets by transfer learning, but as we mentioned, there would be severe negative transfer when using irrelevant facial information (\emph{hair color}).
Our methodology is motivated by the predicament, which is to utilize vast samples that has task-irrelevant features to tackle this problem.

In this paper, we propose \sysName{} to explore task-irrelevant features from large-scale and easily available datasets with the same content as the training set.
By minimizing the Wasserstein distance between the distribution of representation extracted from samples with a fixed task-irrelevant label and that from the entire dataset, we reduce the influence of task-irrelevant features. This suppression of task-irrelevant features plays a ``two negatives make a positive" role that further highlights the representation ability of task-relevant features and finally improve model generalization. 
In summary, in this paper we make three-fold contributions:
\begin{itemize}
    \item We theoretically prove that task-irrelevant labels can help to extract helpful features while only small-scale training dataset is available.
    \item We propose \sysName{}, which use task-irrelevant labels to facilitate the learning process, finally derive a purified feature representation that has minimum task-irrelevant components.
    \item We conduct extensive experiments and analyses on FER and digit recognition. Results show that we achieve state-of-the-art performance on digit recognition task when the training set is small.
\end{itemize}

\section{Theory}
\label{sec:theory}
\subsection{Unreliability of Minimizing Empirical Risk}

For a learning task $\mathcal{T}$, a training sample set including $I$ labeled instances $D_{train} = \{(x_1, y_1), (x_2, y_2), ...(x_I, y_I)\}$ is given. Let $p(x,y)$ be the ground-truth joint probability distribution of sample $x$ and label $y$. We denote the space of input instance by $\mathcal{X}$, and the space of task labels by $\mathcal{Y}$, $x \in \mathcal{X}$ and $y \in \mathcal{Y}$. The goal of deep learning is to learn a model $f(x;\theta):\{\mathcal{X};\Theta\} \rightarrow \mathcal{Y}$ parameterized by $\theta \in \Theta$ from training data to minimize the expected risk, i.e.,
\begin{equation}
\min_{f}R(f)=\int \mathcal{L}(f(x;\theta), y) d p(x, y),
\end{equation}
where $\mathcal{L}$ refers to a certain loss function, e.g., mean squared error or cross entropy loss.

As $p(x,y)$ is unknown, the most classical approach is to approximate the expected risk by minimizing the empirical risk \cite{DBLP:conf/nips/Vapnik91, DBLP:books/daglib/0034861}, which is the average of sample losses over the training set $D_{train}$ of $I$ samples:
\begin{equation}
  \begin{aligned}
    \min _{f} R_{e m p}(f)=\frac{1}{I} \sum_{i=1}^{I} \mathcal{L}\left(y_{i}, f\left(x_{i}\right)\right), \text { s.t. } (x_i, y_i) \in D_{train}.
  \end{aligned}
\end{equation}

However, obviously that when the distribution of the training sample differs greatly from the true distribution $p(x, y)$, especially when $I$ is small, the empirical risk $R_{emp}(f)$ may then be far from being a good approximation of the expected risk $R(f)$, which makes it no longer reliable \cite{3}. The bias of task-irrelevant features in the training dataset is one of the main causes of the deviation between training data distribution and $p(x,y)$. 
We believe that weakening the impact of task-irrelevant features will help train models with better generalization, which requires introducing knowledge of task-irrelevant features from samples with task-irrelevant labels.

\subsection{Task-Irrelevant Labels}
\label{subsec:til}
Task-irrelevant features are defined as the features irrelevant to the target task, and task-relevant features are features relevant to the target task. For example, hair color is a task-irrelevant feature for the FER task. Correspondingly, if samples from some irrelevant tasks contain the same content but different labels with the target task, the labels are termed as task-irrelevant labels, and the samples are termed as task-irrelevant samples. Note that in practice, the task irrelevance of a label is related to the performance loss of transfer learning, not by human judgments. The more negative the transferring effect is, the more irrelevant the label is.

For a specific target task, and a dataset with task-irrelevant labels, $S=\{(x,y_{tir})\}$, we denote the feature space by $X$, $x \in X$, and the label space of task-irrelevant labels by $Y$, $Y=\{y_1,y_2,y_3...y_n\}$, $y_{tir} \in Y$.
Input $x$ contains some features that are task-relevant, denoted by $x_{tr}$ and the distribution of  $x_{tr}$ in $S$ is denoted by $\mathbb{S}$.

We select all samples $S_{y_1}$ whose $y_{tir}=y_1$ from $S$:
\begin{equation}
S_{y_1 } = \{(x,y_{tir})|y_{tir} = y_1,(x,y_{tir})\in S\},
\end{equation}
and the distribution of $x_{tr}$ in $S_{y_1 }$ is denoted by $\mathbb{S}_{y_1}$.
Since $y_{tir}$ is irrelevant with $x_{tr}$, we have:
\begin{equation}
\mathbb{S}= \mathbb{S}_{y_1}.
\end{equation}

In deep models, the feature extractor $f_e$ is used to extract representation $r$ from input data. Here, $R$ and $R_{y_1 }$ are extracted from $S$ and $S_{y_1 }$,
\begin{equation}
\begin{split}
R&=\{(r,y_{tir})| r=f_e(x),(x,y_{tir}) \in S \} \\
R_{y_1 }&=\{(r,y_{tir})| r=f_e(x),(x,y_{tir}) \in S_{y_1 } \}.
\end{split}
\end{equation}
Similarly, the distributions of $r$ in $R$ and $R_{y_1 }$ are denoted by $\mathbb{R}$ and $\mathbb{R}_{y_1 }$.

Under ideal conditions, a good extractor only extracts task-relevant features, so $\mathbb{R}= \mathbb{R}_{y_1 }$. However, in reality, the extractor inevitably extracts task-irrelevant features and leads to $\mathbb{R} \neq \mathbb{R}_{y_1 }$. 
The greater the difference between  $\mathbb{R}$ and $\mathbb{R}_{y_1 }$, the more the extractor is influenced by task-irrelevant features. To reduce this difference, \sysName{} aims to minimize the divergence between these two distributions,
\begin{equation}
\min(\operatorname{WD}(\mathbb{R},\mathbb{R}_{y_1 })),
\label{Wasserstein goal}
\end{equation}
$\operatorname{WD}$ is Wasserstein distance that is used to measure the divergence between two distributions. 

\subsection{Theoretical Analysis of \sysName{}}
 Motivated by \cite{42}, we provide a theoretical analysis of Wasserstein distance's efficacy and its generalization bound here.
The classification model is divided into two parts, a classifier $f_c$, and a feature extractor $f_e$. After sending an input instance to the extractor $f_e$, we get a representation $r$, so $f_e : \mathcal{X} \rightarrow \mathcal{R}$ where $\mathcal{R}$ denotes and $f_c : \mathcal{R} \rightarrow \mathcal{Y}$ where $\mathcal{Y}$
is the space of labels as defined before.

$H$ is a hypothesis class that for every $h \in H$, $h: \mathcal{R} \rightarrow \mathcal{Y}$ and $h$ is K-Lipschitz continuous. In neural networks, we limit the scale of the weights so that $f_c$ is $K$-Lipschitz continuous and $f_c \in H$. 
For every distribution $\mathbb{D}$ on $\mathcal{X}$, the corresponding distribution of representation is denoted by $\mathbb{R}$, $\mathbb{R} = f_e(\mathbb{D})$. 
The difference of $h_1$ and $h_2$ ($h_1,h_2 \in H$) on $\mathbb{R}$ are defined by: 
\begin{equation}
    \epsilon_\mathbb{R}(h_1,h_2) = \underset{r \sim \mathbb{R}}{\mathbb{E}}[||h_1(r)-h_2(r)||].
\end{equation}
Similarly,
\begin{equation}
\epsilon_\mathbb{D}(h_1 \circ f_e,h_2 \circ  f_e) = \underset{x \sim \mathbb{D}}{\mathbb{E}}||h_1(f_e(x))-h_2(f_e(x))||.
\end{equation}

\begin{figure*}[]
\centering
\includegraphics[width=0.95\textwidth]{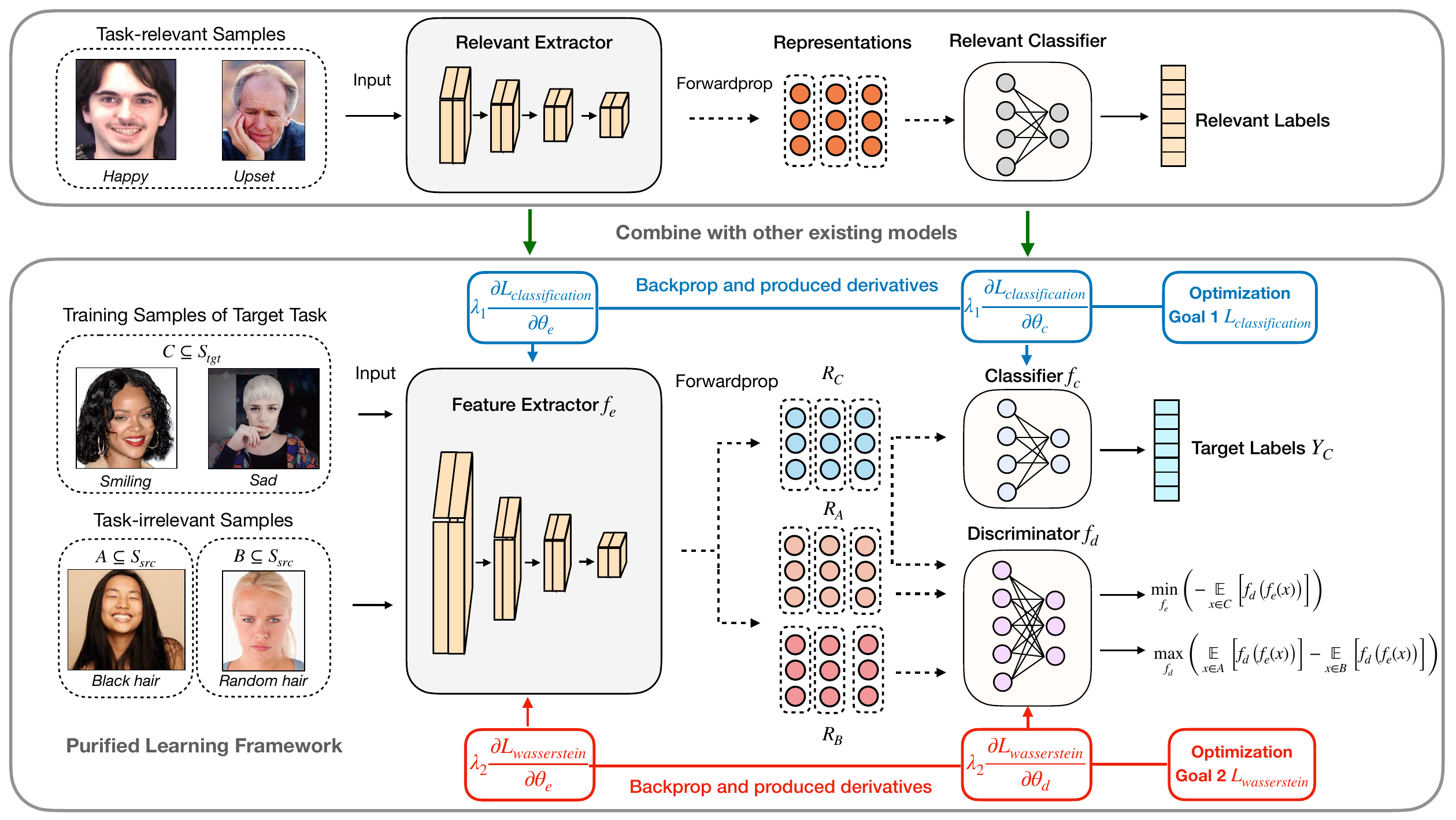}
\caption{An overview of \sysName{} framework, which contains a feature extractor, a discriminator and a classifier. The top of the diagram shows how it can be easily combined with existing models that use task-relevant information.}
\label{framework}
\end{figure*}

$\textbf{Theorem 1}$: For representation distributions $\mathbb{R}_1,\mathbb{R}_2$ on $\mathcal{R}$, and $h_1,h_2 \in H$. Then the following holds:
\begin{equation}
\epsilon_{\mathbb{R}_1}\left(h_1, h_2\right) \leq \epsilon_{\mathbb{R}_2}\left(h_1, h_2\right)+2K \cdot\operatorname{WD} \left(\mathbb{R}_1, \mathbb{R}_2\right).
\end{equation}
\emph{Proof.} We first prove that $\left|h_1-h_2\right|$ is $2K$-Lipschitz continuous. Using the triangle inequality, we have:
\begin{equation}
\begin{aligned}
||h_1(r)-h_2(r)|| \leq&||h_1(r)-h_1(r')||+||h_1(r')-h_2(r)|| \\
 \leq&||h_1(r)-h_1(r')||+||h_1(r')-h_2(r')||\\&+||h_2(r)-h_2(r')||,
\end{aligned}
\end{equation}
thus,
\begin{equation}
\begin{aligned}
&||h_1(r)-h_2(r)|| - ||h_2(r')-h_2(r')|| \\
\leq& ||h_1(r)-h_1(r')|| + ||h_1(r)-h_2(r')||,
\end{aligned}
\end{equation}
then, because both $h_1$ and $h_2$ are $K$-Lipschitz continuous,
\begin{equation} 
\begin{aligned}
&\frac{||h_1(r)-h_2(r)||-||h_1(r')-h_2(r')||}{\rho(r, r')} \\
 \leq& \frac{|h_1(r)-h_1(r')|+||h_2(r)-h_2(r')||}{\rho(r, r')}
 \leq 2 K.
\end{aligned}
\end{equation}
Thus, $\left|h_1-h_2\right|$ is $2K$-Lipschitz continuous. Then we have:
\begin{equation} 
\begin{aligned}
&\epsilon_{\mathbb{R}_1}(h_1, h_2)-\epsilon_{\mathbb{R}_2}(h_1, h_2) 
\\&=\underset{r \in \mathbb{R}_1}{\mathbb{E}}[||h_1(r)-h_2(r)||]
-\underset{r \in \mathbb{R}_2}{\mathbb{E}}[||h_1(r)-h_2(r)||] \\
&\leq \sup _{\|f\|_{L} \leq 2 K} 
\underset{r \sim \mathbb{R}_1}{\mathbb{E}}[f(r)]
-\underset{r \sim \mathbb{R}_2}{\mathbb{E}}[f(r)] \\
&=2K \cdot\operatorname{WD} \left(\mathbb{R}_1, \mathbb{R}_2\right).
\end{aligned}
\end{equation}
So far, \textbf{Theorem 1} is proven.

Next, we give the upper bound of generalization error by Wasserstein distance. Denoting the ideal classifier for a specific classification task by $f^*_c$, both $f_c$ and $f^*_c$ are $K$-Lipschitz continuous. To achieve the optimal performance, $f_c$ needs to approach the ideal classifier $f^*_c$. So the error of $f_c$ on distribution $\mathbb{R}$ is defined as $\gamma_{\mathbb{R}}(f_c)$:
\begin{equation}
\gamma_{\mathbb{R}}(f_c)=\epsilon_{\mathbb{R}}(f_c,f^*_c).
\end{equation}
For two distribution of input instances , denoted by $\mathbb{D}_1$ and $\mathbb{D}_2$, their corresponding representation distributions are $\mathbb{R}_1$ and $\mathbb{R}_2$
According to Theorem 1, we have:
\begin{equation}
\epsilon_{\mathbb{R}_1}(f_c,f^*_c) \leq \epsilon_{\mathbb{R}_2}(f_c,f^*_c) + 2K \cdot\operatorname{WD} \left(\mathbb{R}_1, \mathbb{R}_2\right),
\end{equation}
thus,
\begin{equation}
\gamma_{\mathbb{R}_1}(f_c) \leq \gamma_{\mathbb{R}_2}(f_c) + 2K \cdot\operatorname{WD} \left(\mathbb{R}_1, \mathbb{R}_2\right).
\end{equation}
Correspondingly,
\begin{equation}
\gamma_{\mathbb{D}_1}(f_c  \circ  f_e) \leq \gamma_{\mathbb{D}_2}(f_c \circ  f_e) + 2K \cdot\operatorname{WD} (\mathbb{R}_1, \mathbb{R}_2).
\end{equation}
$\operatorname{WD}$ is Wasserstein distance that is used to measure the divergence between two distributions. Thus, for an unknown test distribution $\mathbb{D}_1$, minimizing error on $\mathbb{D}_1$ can be divided into two goals. The first one is to minimize error on the given training sample distribution $\mathbb{D}_2$. The second is to minimize the Wasserstein distance between $\mathbb{R}_1$ and $\mathbb{R}_2$. 
Since $\mathbb{R}_1$ is unknown, we need to approximate it using another dataset. According to unreliability of empirical risk, we need another large-scale dataset to approximate the distribution of $\mathbb{R}_1$. Large-scale data ensures that the distribution is more close to the testing set. In practice, we use the feature distribution $R$ of $S$, mentioned in Sec. \ref{subsec:til} to do the approximation.

\section{Framework of \sysName{}}

\sysName{} focuses on exploiting the knowledge contained in task-irrelevant labels from auxiliary samples and transferring it to the target task.
With this knowledge, we measure the influence of task-irrelevant features, and then use the adversarial learning method based on Wasserstein distances to limit such influence during feature extraction. In this way, original extract features is purified, increasing the percentage of task-relevant features.
% \subsection{Framework of \sysName{}}
Generally, the framework is divided into three parts: a feature extractor $f_e$, a linear classifier $f_c$ and an additional discriminator $f_d$, as shown in Figure \ref{framework}.  We denote the dataset of the target task by $\mathcal{S}_{tgt}$, and the task-irrelevant dataset by $\mathcal{S}_{src}$. Note that \sysName{} is a theoretically validation framework, the extractor, discriminator and classifier are all flexible and have multiple choices. A practice choice is to set them according to performance. 
To utilize both $\mathcal{S}_{tgt}$ and $\mathcal{S}_{src}$, two optimization goals are set according to Sec. \ref{sec:theory}. 

\subsection{Goal 1: Empirical Risk Minimization}
For training sample $x$ and label $y$, $(x,y) \in \mathcal{S}_{tgt}$, the classification probability $p = \operatorname{softmax}(f_c(f_e(x)))$,
and the classification loss is calculated by cross-entropy loss, which is consistent with the loss in empirical risk.
\begin{equation}
\begin{split}
{\mathcal{L}_{\text {classification}}}=& \\
- \sum_{x,y \in \mathcal{S}_{tgt}}& y * \log \left(\operatorname{softmax}\left(f_{c}(f_e(x))\right)\right).
\end{split}
\end{equation}

Goal 1 is the empirical risk minimization that we minimize the classification error so that the model learns the knowledge contained in the samples of $\mathcal{S}_{tgt}$. However, the generalization ability of the model trained on Goal 1 depends heavily on the consistency of the training data distribution and the real data distribution as we discussed before, which is usually hard to achieve on a small dataset. Thus, a new optimization goal should be imported to utilize large-scale task-relevant or task-irrelevant data to reduce this bias.

\subsection{Goal 2: Wasserstein Distance Minimization}

This goal aims to minimize the Wasserstein distance between $\mathbb{R}_{y_1 }$ and $\mathbb{R}$ as Equation \ref{Wasserstein goal}, and it is necessary to estimate the two distributions by sampling. 
For $\mathcal{S}_{src}$, we randomly select two groups of samples, one from samples with a specific task-irrelevant label $y_1$ in $\mathcal{S}_{src}$, denoted by $A$, and the other from the entire $\mathcal{S}_{src}$, denoted by $B$. For $A$ and $B$, the representations extracted by $f_e$ are denoted by $R_A$, $R_B$:
\begin{equation}
\begin{split}
R_A =& \{ f_e(x) | x \in A \} \\
R_B =& \{ f_e(x) | x \in B \}.
\end{split}
\end{equation}

% %%%%%%%%%%%%%%%%%%%%%%%%%%%%%%%%%%%%%%%%%%%%%%%%%%%%%%%%%%%%%%%%%%%%%%%%%%
\begin{algorithm}[t]
\small
    \caption{Algorithm of \sysName{}.}
    \label{algo}
    \hspace*{0.05in} {\bf Require:}
  target training sample set $\mathcal{S}_{tgt}$; task-irrelevant sample set $\mathcal{S}_{src}$; learning rate for discriminator $\alpha_1$; learning rate for feature extractor and classifier $\alpha_2$; batch size $m$; Iteration numbers $n_1$, $n_2$. \\
    \hspace*{0.05in} {\bf Notation:} 
  function fitted by the feature extractor, $f_e$; function fitted by the classifier, $f_c$; function fitted by the discriminator, $f_d$.
    \begin{algorithmic}[1]
        \State Initialize feature extractor, linear classifier, discriminator with random weights $\theta_e$ , $\theta_c$ , $\theta_d$. 
        \For{$n_1$ steps} 
        \State Sample minibatch $A = \{a^{(i)}\}_{i=1}^m$ from $\mathcal{S}_{src}$ with a fixed task-irrelevant label.
        \State Sample minibatch $ B=\{b^{(i)}\}_{i=1}^m$ from $\mathcal{S}_{src}$ with random task-irrelevant labels.
        \State $g_d \leftarrow \bigtriangledown_d (\frac{1}{m} \sum^m_{i=1} (f_d(f_e(b^{(i)})) - f_d(f_e(a^{(i)})))$
        \State $\theta_d \leftarrow \theta_d + \alpha_1 Adam(\theta_d, g_d)$
        \For{$n_2$ steps}
        \State Sample minibatch $C=\{x^{(i)}, y^{(i)}\}_{i=1}^m$ from $\mathcal{S}_{tgt}$.
        \State \textbf{Goal 1}:
        \State $g_c \leftarrow \bigtriangledown_c (\sum^m_{i=1} (-y^{(i)} \cdot log(f_c(f_e(x^{(i)})))))$ 
        \State $\theta_c \leftarrow \theta_c - \alpha_2 SGD(\theta_c, g_c)$
        \State $g_e \leftarrow \bigtriangledown_e (\sum^m_{i=1} (-y^{(i)} \cdot log(f_c(f_e(x^{(i)})))))$ 
        \State $\theta_e \leftarrow \theta_e - \alpha_2 SGD(\theta_e, g_e)$
        %\State $g_e,g_c \leftarrow \bigtriangledown_{e,c} (\sum^m_{i=1} (-y^{(i)} \cdot log(f_c(f_e(x^{(i)})))))$ 
        %\State $\theta_e,\theta_c \leftarrow \theta_e - \alpha_2 SGD(\theta_e, g_e),\theta_c - \alpha_2 SGD(\theta_c, g_c)$
        \State \textbf{Goal 2}:
        %\State $g_e \leftarrow \bigtriangledown_e (\sum_{i=1}^m (-y^{(i)} log(f_c(f_e(x^{(i)})))) + \frac{1}{m} \sum^m_{i=1} (-f_d(f_e(x^{(i)})))$
        \State $g_e \leftarrow \bigtriangledown_e (\frac{1}{m} \sum^m_{i=1} (-f_d(f_e(x^{(i)}))))$
        \State $\theta_e \leftarrow \theta_e - \alpha_2 SGD(\theta_e, g_e)$ 
        \EndFor
        \EndFor
    \end{algorithmic}
\end{algorithm}
% %%%%%%%%%%%%%%%%%%%%%%%%%%%%%%%%%%%%%%%%%%%%%%%%%%%%%%%%%%%%%%%%%%%%%%%%%%

When the sizes of $A$ and $B$ are large enough, the representation distributions of $R_A$ and $R_B$, denoted by $\mathbb{R}_A$ and $\mathbb{R}_B$, are used as reasonable estimations of $\mathbb{R}_{y_1}$ and $\mathbb{R}$.
Thus, Equation \ref{Wasserstein goal} is rewritten as:
\begin{equation}
\min(\operatorname{WD}(\mathbb{R}_A,\mathbb{R}_B)).
\end{equation}
We substitute the $\operatorname{WD}$ with Wasserstein's Kantorovich-Rubinstein duality form,
\begin{equation}
\min_{f_e}\sup _{\left\|f_d\right\|_{L \leq 1}}\left(\underset{x \in A}{\mathbb{E}}\left[f_d(f_e(x))\right]-\underset{x \in B}{\mathbb{E}}\left[f_d(f_e(x))\right]\right).
\label{goal real}
\end{equation}

Inspired by WGAN \cite{41}, Equation \ref{goal real} could be divided into two steps. Firstly, the discriminator $f_d$ is trained by:
\begin{equation}
\max _{f_d}\left(\underset{x \in A}{\mathbb{E}}\left[f_d(f_e(x))\right]-\underset{x \in B}{\mathbb{E}}\left[f_d(f_e(x))\right]\right).
\end{equation}
Secondly, we select a group of training samples, $C$, from training dataset of target task $\mathcal{S}_{tgt}$. The feature representations extracted by $f_e$ are denoted by $R_C$,
\begin{equation}
R_C=  \{ f_e(x) |  x \in C \},
\end{equation}
then, the extractor $f_e$ is trained by:
\begin{equation}
\min_{f_e}\left(-\underset{x \in C}{\mathbb{E}}\left[f_d(f_e(x))\right]\right),
\end{equation}
and the corresponding Wasserstein loss is:
\begin{equation}
{\mathcal{L}}_{\text {wasserstein}}=-\frac{1}{\|C\|}  \underset{x \in C}{\sum}\left[f_d(f_e(x))\right],
\end{equation}
where $\|C\|$ is the size of the sample group C.

Combining $\mathcal{L}_{\text {classification}}$ and $\mathcal{L}_{\text {wasserstein}}$, the complete loss function is written as follows, where $\lambda_i$ are weighting factors:
\begin{equation}
\mathcal{L}_{\text{Purify Learning}}=\lambda_{1} {\mathcal{L}}_{\text {classification}}+\lambda_{2}  {\mathcal{L}}_{\text {wasserstein}}.
\end{equation}

In practice, the training algorithm is shown in Alg. \ref{algo}.

\section{Experiments}

% Table 1
\begin{table*}[h]
\centering
\scalebox{1.00}{
\begin{tabular}{ccccccc}
\toprule
Feature Extractor & Train $\rightarrow$ Test Dataset & Goal 1 Only & \begin{tabular}[c]{@{}c@{}}TL\\ (Hair Color)\end{tabular} & \begin{tabular}[c]{@{}c@{}}MTL\\ (Hair Color)\end{tabular} & \begin{tabular}[c]{@{}c@{}}AMTL\\ (Hair Color)\end{tabular} & \begin{tabular}[c]{@{}c@{}}PL \\ (Hair Color)\end{tabular} \\ \midrule

\multirow{7}{*}{AlexNet} & ck+ $\rightarrow$ mmi & 35.91 & \textbf{38.28} & 33.73  & 34.57 & 37.61 \\
        & ck+ $\rightarrow$ oulu & 34.75 & 25.54 & 33.08 & 33.15 & \textbf{38.31} \\
        & mmi $\rightarrow$ ck+  & 56.36 & 44.97 & 57.09 & 57.94 & \textbf{61.45} \\
        & mmi $\rightarrow$ oulu & 22.12 & 19.89 & 23.66 & \textbf{28.73} & 22.61 \\
        & oulu $\rightarrow$ ck+ & 55.03 & 54.42 & 54.67 & 43.52 & \textbf{59.64} \\
        & oulu $\rightarrow$ mmi & 39.46 & 40.00 & 40.30 & 28.33 & \textbf{45.36} \\
        & Average                & 40.61 & 37.18 & 40.42 & 37.37 & \textbf{44.16} \\ 
\midrule
\multirow{7}{*}{ResNet34} & ck+ $\rightarrow$ mmi &\textbf{50.42} & 46.54 & 46.54 & 44.35 & \textbf{50.42}  \\
        & ck+ $\rightarrow$ oulu& 50.94 &\textbf{54.29} & 47.24 & 43.75 & 51.78  \\
        & mmi $\rightarrow$ ck+ & 65.33 & 68.73 & 66.18 & 64.61 & \textbf{69.21} \\
        & mmi $\rightarrow$ oulu & 44.87& \textbf{46.13} & 40.96 & 42.15 & 42.43 \\
        & oulu $\rightarrow$ ck+ & 73.45& 73.33 & 72.93 & 66.18 & \textbf{80.12} \\
        & oulu $\rightarrow$ mmi & 54.13& 49.01 & 51.43 & 41.99 & \textbf{54.30} \\
        & Average & 56.52& 56.34 & 54.21 & 50.51 & \textbf{58.04}          \\ 
\midrule
\multirow{7}{*}{VggNet19}  & ck+ $\rightarrow$ mmi       & 45.53& 41.10                   & 37.10                     & 33.05                                   & \textbf{45.53}          \\
          & ck+ $\rightarrow$ oulu      & 56.66& 44.52                   & 32.24                     & 47.66                                   & \textbf{57.22}          \\
          & mmi $\rightarrow$ ck+       & 65.33& 62.30                   & 64.61                     & 57.82                                   & \textbf{66.79}          \\
          & mmi $\rightarrow$ oulu      & 45.08& 32.17                   & 46.69                     & 40.20                                   & \textbf{46.76}          \\
          & oulu $\rightarrow$ ck+      & 76.73& 71.27                   & 72.12                     & 31.88                                   & \textbf{78.91}          \\
          & oulu $\rightarrow$ mmi      & 45.03& 43.68                   & 40.00                     & 20.24                                   & \textbf{51.43}          \\
          & Average                     & 55.73& 49.17                   & 48.79                     & 38.48                                   & \textbf{57.77}          \\ 
\bottomrule
\end{tabular}
}
\caption{Results (Accuracy $\%$) on FER task, where TL refers to transfer learning, MTL refers to multi-task learning, AMTL refers to adversarial multi-task learning, and PL refers to \sysName{}. ck+$\rightarrow$mmi means model is trained on ck+ and tested on mmi dataset. TL (Hair Color) means using \emph{hair color} label in transfer learning. }
\label{table:fer}
% \vspace{-1em}
\end{table*}

\subsection{Experimental Setup}

\subsubsection{Model}
As mentioned before, the framework of \sysName{} consists of three parts: a feature extractor, a linear classifier, and a discriminator. The discriminator consists of four fully connected layers. The linear classifier is a fully connected layer with the number of neurons depending on the specific task. The flexibity of \sysName{} enables that various mainstream models can be embed as feature extractor.

\subsubsection{Evaluation Metrics}
To evaluate the results of model generalization, the cross-dataset test results are taken as evaluation metrics. 
This is based on a reasonable assumption: the training set and the real data distribution are different, and the unseen test set is sampled from the real data distribution, so there exists a gap between the distribution of test data and that of training data. Therefore, the higher the accuracy in our test means the better the generalization performance of the model.

\subsubsection{Hyperparameters}
We implement all experiments without data augmentation. The model is trained by an SGD optimizer with an initial learning rate of 0.001, the momentum of 0.9, StepLR(step size is 7), and $\gamma$ of 0.1. 
We apply widely used classification models (e.g ResNet) as a feature extractor and set the output dimension of the penultimate layer to 128. The classifier is a fully connected layer and the output dimension is equal to the number of classes (7 for FER and 10 for digit recognition). The discriminator consists of four fully connected layers with 512, 256, 10, 1 nodes, the discriminator is trained using an Adam with the same learning rate, and a weight limit 0.1. In addition, the step ratio of  $n_2$ in Alg.1 is 3. The factors $\lambda_1$ and $\lambda_2$ are both 1. We use batch size 32 in FER, and 128 in digit recognition. All results are obtained after the model has been trained for 50 epochs.

\subsection{Experimental implementation details} 
\label{experimental_details}
We provide implementation details of the comparison method used in our experiments.
\subsubsection{Facial Expression Recognition}
\begin{itemize}
\item \textbf{Transfer learning}: We adopt the pre-training/fine-tuning approaches to transfer learning based on shared parameters \cite{yosinski2014transferable}. We first pre-train the feature extractor as well as the classifier on CelebA for hair color classification and then train the whole model on the expression dataset.
\item \textbf{Multi-task learning}: We adopt the multi-task learning approach with hard parameter sharing of hidden layers \cite{54}. Specifically, a shared feature extractor accepts both images with hair color tags from CelebA and images with expression tags from a training set. And two independent classifier networks make separate predictions for hair color and expressions. The entire network gets trained by backpropagating the final loss which is calculated by adding up the classification loss of hair color and the classification loss of expression.
\item \textbf{Adversarial Multi-task Learning}: Based on the multi-tasking learning model, a new gradient reversal layer (GRL) is added between the feature extractor and the hair color classifier, where the idea is consistent with Shinohara \cite{14}.
\end{itemize}

\subsubsection{Digit Recognition}
\begin{itemize}
\item \textbf{DANN}: DANN is a framework for unsupervised domain adaptation based on adversarial multitasking learning proposed by Ganin \cite{mnist-m}. As a comparison method for the simultaneous use of MNIST and SVHN data, it is implemented by adding a new gradient reversal layer (GRL) and a domain classifier with 2-dimensional outputs for domain classification to the existing feature extractor and classifier. The feature extractor accepts both images from MNIST and SVHN. One classifier then classifies the feature representation of the MNIST data from 0 to 9, and another classifier distinguishes whether the feature representation originates from MNIST or SVHN.
\end{itemize}

\subsection{Experiments on Facial Expression Recognition}
In this part, we compare \sysName{} with some methods related to our work on the FER task. These methods are briefly described below, and their implementation details are shown in Section \ref{experimental_details}:
\begin{itemize}[noitemsep]
\item \textbf{Goal 1 Only}: %only use the optimization goal 1.
As an empirical baseline, we use labeled training data to train the model with the optimization goal 1 only.
\item \textbf{Transfer learning }: We use the hair color recognition task to pre-train the network, and then fine-tune on FER task.
\item \textbf{Multi-task learning }: We train FER and hair color recognition tasks using a shared feature extractor.
\item \textbf{Adversarial multi-task learning}: We add a gradient inversion layer for the hair color recognition task on the multi-task learning method above.
%Based on the multi-task learning method, a gradient inversion layer is added for the hair color recognition task.
\end{itemize}

\subsubsection{Datasets and Feature Extractors}
In FER experiments, we use small-scale datasets, including ck+  \cite{ck}, oulu \cite{oulu} and mmi \cite{mmi} for cross-dataset evaluation. We select AlexNet \cite{alexnet}, VggNet19  \cite{vggnet} and ResNet34 \cite{resnet} as feature extractors.

%%%%%%%%%%%%%%%%%%%%%%%%%%%%%%%%%%%%%%%%%%%%%%%%%%%%%%%%%%%%%%%%%%%%%%%%%%%%%%%%%%%%%%%%%%%%%%%%%%%%%%%%%%%%%%%%%%%%
We use $224\times224$ resolution for all RGB pictures and preprocess them through MTCNN \cite{mtcnn} for alignment. More importantly, the \emph{hair color} label in CelebA \cite{celeba} is regarded as task-irrelevant label. Therefore, we select samples with fixed \emph{hair color} labels and samples randomly selected from CelebA, and then using 
\sysName{} on them.

\subsubsection{Results}
Table \ref{table:fer} is the comparison between \sysName{} and baseline methods, which shows that our method achieves better results in 14 of 18 cross-dataset tests.

Since hair color is not relevant to FER, it leads to negative transfer for transfer learning methods including transfer learning, multi-task learning and adversarial multi-task learning (lower accuracy even than Goal 1 Only in most cases). The performance of the adversarial multi-task learning is the worst because it requires training samples with both task label and task-irrelevant label. However, in our experiments, the training samples are only annotated with task labels, and the task-irrelevant labels are from additional samples. By contrast, \sysName{} is able to make use of the knowledge from task-irrelevant labels, so it improves the accuracy of Goal 1 only, thus successfully resolving negative transfer and improving the generalization performance of the model.

\subsection{Experiments on Digit Recognition}

In this part, we evaluate \sysName{} on Digit Recognition task. In this experiment, we add another baseline, DANN \cite{45}, a domain transfer framework based on adversarial multi-task learning. DANN is currently the state-of-the-art method of transferring from MNIST to MNIST-M\footnote{\href{https://paperswithcode.com/sota/domain-adaptation-on-mnist-to-mnist-m}{Link to Paperwithcode}}. Since we use small-scale training set, our accuracy is lower than original paper.
The model details of DANN have been shown in Section \ref{experimental_details}.

\subsubsection{Datasets and Feature Extractors}
In the digit recognition experiments, regarding \emph{background color} as the task-irrelevant features, we also use a small-scale training set, which are 20000 pictures selected from MNIST \cite{mnist}. The test sets are SVHN \cite{svhn} and MNIST-M \cite{mnist-m}. The two groups of samples are randomly sampled from MNIST and the combination of MNIST and SVHN respectively. Besides, all pictures are converted to RGB pictures and we use $32\times32$ resolution.

Additionally, we select ResNet18 \cite{resnet}, VggNet11 \cite{vggnet}, DenseNet121 \cite{densenet}, SeNet \cite{senet}, EﬃcientNet \cite{efficientnet}, MobileNetV2 \cite{mobilenetv2} and ShuﬄeNetV2 \cite{shufflenetv2} as feature extractors.

\subsubsection{Results}

\begin{table}[htb]
%\resizebox{80mm}{25mm}{
\centering
\scalebox{0.90}{
\begin{tabular}{ccccc}
\toprule
\begin{tabular}[c]{@{}c@{}}Train $\rightarrow$ Test\\ Dataset\end{tabular} & 
\begin{tabular}[c]{@{}c@{}}Feature\\ Extractor \end{tabular} & 
\begin{tabular}[c]{@{}c@{}}Goal 1\\ Only\end{tabular} & 
DANN &
\begin{tabular}[c]{@{}c@{}}PL \\ (Background Color)\end{tabular}              \\ 
\midrule
\multirow{7}{*}{\begin{tabular}[c]{@{}c@{}}MNIST $\rightarrow$ \\ SVHN\end{tabular}}                                  & DenseNet121                            & 13.85                           & 10.64                    & \textbf{16.44} \\
                                 & EfficientNet                           & 22.99                           & 13.99                    & \textbf{33.74} \\
                                 & MobileNetV2                            & 20.06                           & 11.35                    & \textbf{22.11} \\

                                 & ResNet18                               & 15.01                           & 12.00                    & \textbf{18.83} \\
                                 & SeNet                                  & 15.44                           & 10.69                    & \textbf{18.22} \\
                                 & ShuffleNetV2                           & 14.29                           & 11.98                    & \textbf{20.72} \\
                                 & VggNet11                               & 19.50                           & 13.79                    & \textbf{25.71} \\ 
%\cline{2-5} 
                                 & Average                                & 17.31                           & 12.06                    & \textbf{22.25} \\ 
\midrule
\multirow{7}{*}{\begin{tabular}[c]{@{}c@{}}MNIST $\rightarrow$ \\ MNIST-M\end{tabular}}                                  & DenseNet121                            & 25.26                           & 14.28                    & \textbf{35.21} \\
                                 & EfficientNet                           & 40.78                           & 38.71                    & \textbf{57.64} \\
                                 & MobileNetV2                            & 29.77                           & 21.75                    & \textbf{48.76} \\
                                                                                                          & ResNet18                               & 25.10                           & 22.54                                                     & \textbf{37.64} \\
                                 & SeNet                                  & 17.24                           & 15.83                    & \textbf{39.71} \\
                                 & ShuffleNetV2                           & 27.96                           & 20.34                    & \textbf{47.65} \\
                                 & VggNet11                               & 39.18                           & 42.09                    & \textbf{48.98} \\ 
%\cline{2-5} 
                                 & Average                                & 29.33                           & 25.08                    & \textbf{45.08} \\ 
\bottomrule
\end{tabular}
}
\caption{Results (Accuracy $\%$) on digit recognition task.}
\label{table:dr}
% \vspace{-1.5em}
\end{table}

Table \ref{table:dr} is the comparison between our method and two baselines. Compared with baselines, \sysName{} achieves higher accuracy than DANN and Goal 1 Only method. The 
Although transferring from MNIST to SVHN is difficult since SVHN is much more complex than MNIST, \sysName{} improves DANN by more than 10\% accuracy, which proves the effectiveness of using task-irrelevant features. We also achieve better results than DANN in the “MNIST to MNIST-M” task.

\subsection{Discussion}

\subsubsection{The impact of different levels of task-irrelevance.}

%table 3
\begin{table}[htb]
\centering

\scalebox{0.9}{
\begin{tabular}{ccccc}
\toprule
\begin{tabular}[c]{@{}c@{}}Feature\\ Extractor\end{tabular} & \begin{tabular}[c]{@{}c@{}}Train $\rightarrow$ Test\\ Dataset\end{tabular} & \begin{tabular}[c]{@{}c@{}}Goal 1\\ Only\end{tabular} & \begin{tabular}[c]{@{}c@{}}PL\\ (Smiling)\end{tabular} & \begin{tabular}[c]{@{}c@{}}PL\\ (Hair Color)\end{tabular} \\ 
\midrule
\multirow{7}{*}{AlexNet} & ck+ $\rightarrow$ mmi & 35.91 & 32.38 & \textbf{37.61} \\
 & ck+ $\rightarrow$ oulu & 34.75 & 33.91 & \textbf{38.31} \\
 & mmi $\rightarrow$ ck+ & 56.36 & 35.39 & \textbf{61.45} \\
 & mmi $\rightarrow$ oulu & 22.12 & 18.77 & \textbf{22.61} \\
 & oulu $\rightarrow$ ck+ & 55.03 & 49.33 & \textbf{59.64} \\
 & oulu $\rightarrow$ mmi & 39.46 & 38.45 & \textbf{45.36} \\
 & Average & 40.61 & 34.71 & \textbf{44.16} \\ 
\midrule
\multirow{7}{*}{ResNet34} & ck+ $\rightarrow$ mmi & \textbf{50.42} & 47.89 & \textbf{50.42} \\
 & ck+ $\rightarrow$ oulu & 50.94 & 50.31 & \textbf{51.78} \\
 & mmi $\rightarrow$ ck+ & 65.33 & 65.70 & \textbf{69.21} \\
 & mmi $\rightarrow$ oulu & \textbf{44.87} & 39.22 & 42.43 \\
 & oulu $\rightarrow$ ck+ & 73.45 & 73.70 & \textbf{80.12} \\
 & oulu $\rightarrow$ mmi & 54.13 & 51.43 & \textbf{54.30} \\
 & Average & 56.52 & 54.71 & \textbf{58.04} \\ 
\midrule
\multirow{7}{*}{VggNet19} & ck+ $\rightarrow$ mmi & 45.53 & 40.47 & \textbf{45.53} \\
 & ck+ $\rightarrow$ oulu & 56.66 & 42.99 & \textbf{57.22} \\
 & mmi $\rightarrow$ ck+ & 65.33 & 63.64 & \textbf{66.79} \\
 & mmi $\rightarrow$ oulu & 45.08 & 44.59 & \textbf{46.76} \\
 & oulu $\rightarrow$ ck+ & 76.73 & 75.03 & \textbf{78.91} \\
 & oulu $\rightarrow$ mmi & 45.03 & 50.59 & \textbf{51.43} \\
 & Average & 55.73 & 52.89 & \textbf{57.77} \\ 
\toprule
\end{tabular}
} 
\caption{Results (Accuracy $\%$) on FER task, using different labels (\emph{smiling} and \emph{hair color}) in \sysName{}.
}
\label{table:fer-labels}
% \vspace{-1.5em}
\end{table}

As different task-relevance can affect the performance of transfer learning, it is essential to study the impact of different levels of task-irrelevance on \sysName{}.
Except for the \emph{hair color} label, we use the \emph{smiling} label from CelebA. We apply these labels to the \sysName{} framework and compare their performance.

Table \ref{table:fer-labels} indicates that the experiment results that compared with Goal 1 Only method, the performance of \sysName{} remains the same or gets worse when using the \emph{smiling} label, but have significant improvement when using the \emph{hair color} label. 
The experiment results suggest that using a more irrelevant label is better for the \sysName{} framework to improve the performance since it tries to use the knowledge of task-irrelevant features. On the other hand, if we misuse task-relevant samples in \sysName{}, it will have a negative impact. This is similar to negative transfer while using task-irrelevant labels in transfer learning. In general, this phenomenon means that \sysName{} differs from transfer learning, since it benefits from the irrelevance of data instead of relevance.

\subsubsection{PCA analysis on feature representations}
Based on the maximum variance theory \cite{PCA_theory}, if the percentage of variance explained by principal components increase, the representation vector will contain more information about task-relevant features, and task-relevant information will be better retained during the process of dimension reduction.

\begin{figure}[htb]

\centering

\includegraphics[height=5cm,width=8.5cm]{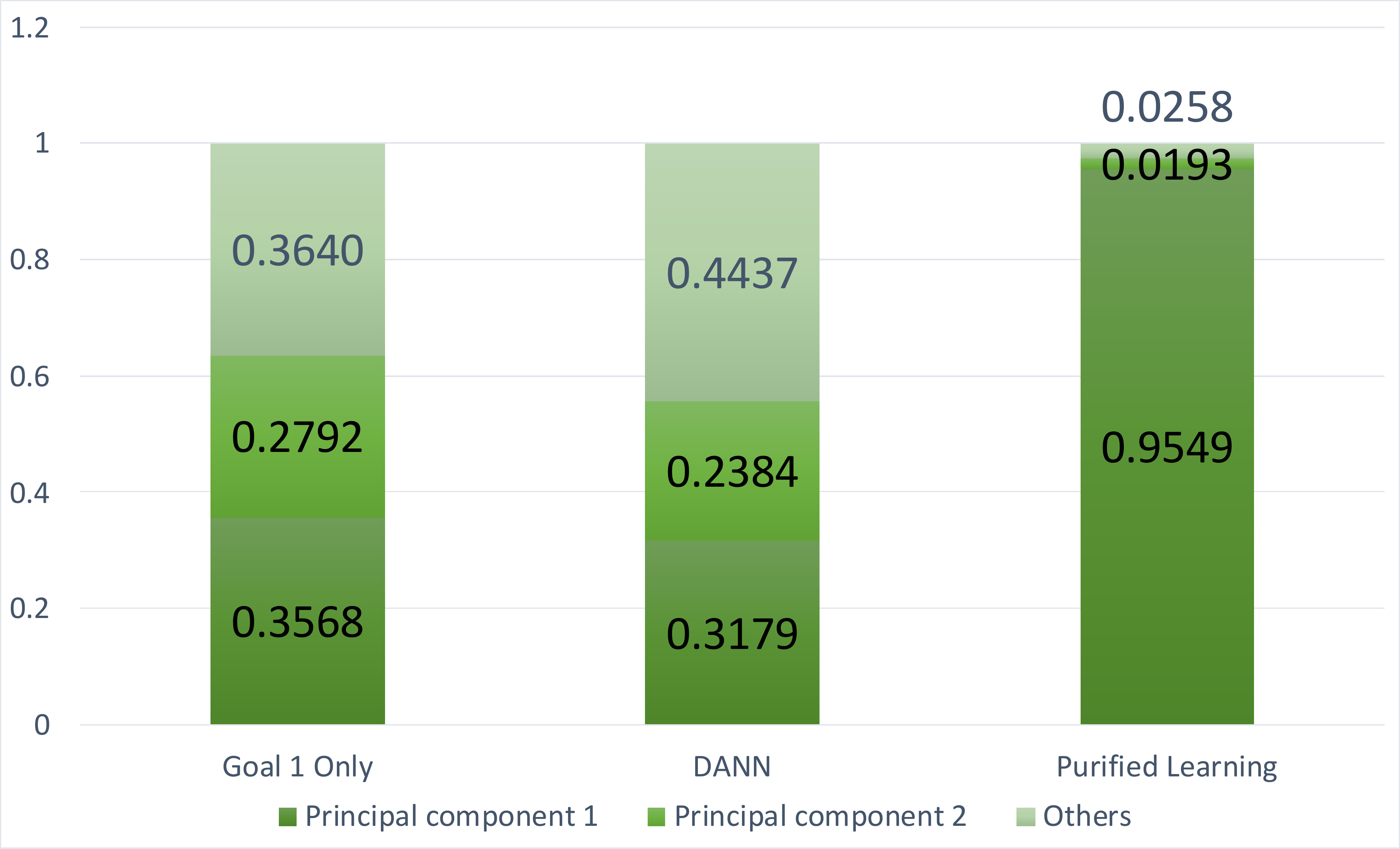}
\caption{PCA results of ResNet18 on digit recognition task. The larger the proportion of the dark green part (principal component 1 $\&$ 2), the larger the proportion of the components that are decisive to the task in the representation.}
\label{PCA}
\end{figure}

Figure \ref{PCA}  shows the proportion of principal components of different methods in digit recognition tasks. Compared with other methods (Goal 1 Only and DANN), the proportion of task-relevant information is significantly higher in \sysName{}, which means that \sysName{} learns knowledge from task-irrelevant labels and avoid extracting task-irrelevant features during feature extraction.

\section{Conclusion}
In this paper, we propose a novel framework, \sysName{}, to exploit the knowledge from additional task-irrelevant labels in order to solve the problem caused by the unreliable empirical risk minimization when the training dataset is small. Based on detailed theoretical analysis, we illustrate that samples with task-irrelevant labels can be used in improving the generalization performance of the model, based on which we propose \sysName{}, which directly obtains and utilizes knowledge from a wide range of task-irrelevant labels. Furthermore, \sysName{} can be well combined with the existing task-relevant learning methods. We believe that the exploration of task-irrelevant labels in our work provides a valuable insight for future research.

\section{Acknowledgement}
This research is supported by National Natural Science Foundation of China (Grant No. 6201101015 and 61602013), Beijing Academy of Artificial Intelligence (BAAI), the Natural Science Foundation of Guangdong Province (Grant No. 2021A1515012640), the Basic Research Fund of Shenzhen City (Grant No. JCYJ20210324120012033 and JCYJ20190813165003837), the Shenzhen General Research Project (Grant No. JCYJ20190808182805919), the National Key R\&D Program of China (No. 2021ZD0112905), the Overseas Cooperation Research Fund of Tsinghua Shenzhen International Graduate School (Grant No. HW2021008) and the 173 program (Grant No. 2021-JCJQ-JJ-0029).

\clearpage

\bibliographystyle{IEEEtranS}
\bibliography{IEEEexample}

\end{document}